\documentclass{article}

\usepackage{microtype}
\usepackage{graphicx}
\usepackage{subfigure}
\usepackage{booktabs} 
\usepackage{multirow}
\usepackage{listings}
\usepackage{xcolor}

\definecolor{codegreen}{rgb}{0,0.6,0}
\definecolor{codegray}{rgb}{0.5,0.5,0.5}
\definecolor{codepurple}{rgb}{0.58,0,0.82}
\definecolor{backcolour}{rgb}{0.95,0.95,0.92}

\lstdefinestyle{mystyle}{
    backgroundcolor=\color{backcolour},   
    commentstyle=\color{codegreen},
    keywordstyle=\color{magenta},
    numberstyle=\tiny\color{codegray},
    stringstyle=\color{codepurple},
    basicstyle=\ttfamily\footnotesize,
    breakatwhitespace=false,         
    breaklines=true,                 
    captionpos=b,                    
    keepspaces=true,                 
    numbers=left,                    
    numbersep=5pt,                  
    showspaces=false,                
    showstringspaces=false,
    showtabs=false,                  
    tabsize=2
}

\usepackage{hyperref}

\usepackage[accepted]{mlsys2022}

\mlsystitlerunning{Graph Neural Network Training with Data Tiering}

\begin{document}

\twocolumn[
\mlsystitle{Graph Neural Network Training with Data Tiering}



\mlsyssetsymbol{equal}{*}

\begin{mlsysauthorlist}
\mlsysauthor{Seung Won Min}{UIUC}
\mlsysauthor{Kun Wu}{UIUC}
\mlsysauthor{Mert Hidayetoğlu}{UIUC}
\mlsysauthor{Jinjun Xiong}{Buffalo}
\mlsysauthor{Xiang Song}{AWS}
\mlsysauthor{Wen-mei Hwu}{NVIDIA,UIUC}
\end{mlsysauthorlist}

\mlsysaffiliation{UIUC}{University of Illinois at Urbana-Champaign}
\mlsysaffiliation{NVIDIA}{NVIDIA}
\mlsysaffiliation{AWS}{AWS Shanghai AI Lab}
\mlsysaffiliation{Buffalo}{University of Buffalo}

\mlsyscorrespondingauthor{Seungwon Min}{min16@illinois.edu}

\mlsyskeywords{Machine Learning, MLSys}

\vskip 0.3in

\begin{abstract}
Graph Neural Networks (GNNs) have shown success in learning from graph-structured data, with applications to fraud detection, recommendation, and knowledge graph reasoning.
However, training GNN efficiently is challenging because: 1) GPU memory capacity is limited
and can be insufficient for large datasets, and 2) the graph-based data structure causes irregular data access patterns.
In this work, we provide a method to statistical analyze and identify more frequently accessed data ahead of GNN training.
Our data tiering method not only utilizes the structure of input graph, but also an insight gained from actual GNN training process to achieve a higher prediction result.
With our data tiering method, we additionally provide a new data placement and access strategy to further minimize the CPU-GPU communication overhead.
We also take into account of multi-GPU GNN training as well and we demonstrate the effectiveness of our strategy in a multi-GPU system.
The evaluation results show that our work reduces CPU-GPU traffic by 87--95\% and improves the training speed of GNN over the existing solutions by 1.6--2.1$\times$ on graphs with hundreds of millions of nodes and billions of edges.

\end{abstract}
]



\printAffiliationsAndNotice{} 

\section{Introduction}
Graph neural networks (GNNs) have shown promising successes on multiple graph-based machine learning tasks including fraud detection~\cite{10.1145/3397271.3401253}, recommendation~\cite{10.1145/3308558.3313488}, search and knowledge graph reasoning~\cite{dettmers2018convolutional}.
With a rapidly growing need to apply GNNs in various domains, there are several efforts from the community to provide open-source GNN-specific machine learning frameworks such as PyTorch Geometric (PyG)~\cite{Fey/Lenssen/2019}, Deep Graph Library (DGL)~\cite{wang2019dgl}, and Spektral~\cite{DBLP:journals/corr/abs-2006-12138}.
Those graph-specific frameworks implement several highly optimized message passing operators and graph-specific computation layers which were lacking in the previous DNN frameworks.

Yet, the challenges of GNN training are not limited to the message passing or the computational layers.
Recently, the training of GNNs has been widening to very large graphs.
With the successes of using large datasets in machine learning to increase the training accuracy~\cite{DBLP:journals/corr/RussakovskyDSKSMHKKBBF14, wu2018moleculenet}, the importance of using larger graphs took a place in GNN training as well~\cite{DBLP:journals/corr/abs-2103-09430}.
The number of nodes and the edges of these graphs reach millions to billions~\cite{ugander2011anatomy,zhu2019aligraph} and the graphs with such scales make the ordinary na\"ive software/hardware approaches ineffective.

The earlier implementations of GNN were mostly focusing on a small scale graphs~\cite{DBLP:journals/corr/KipfW16,velickovic2018graph} and assumed the whole graph 
fits into a single GPU memory.
%
Therefore, previously, accessing an arbitrary node's feature data was merely a process of indexing the GPU's own memory space.
%
%
%
However, for large graphs whose node/edge feature data cannot fit into the GPU memory, at least part of the graph needs to be placed into the CPU memory.
One common practice to train GNNs 
in such scenario is to create a smaller set of problem by performing a mini-batched training.
With the mini-batch training, only a subset of nodes are randomly picked along with their neighboring nodes and sent to GPU.
This method is very effective when training GNNs on large graphs as it practically reduces the memory footprint of the application.
Not only that, several recently introduced GNN models~\cite{Preprint2021RUNIMPSF,DBLP:journals/corr/abs-2107-09422,Preprint2021SYNERISE} showed that the mini-batched based approaches are superior in achieving high training accuracy as well.

A mini-batch training process that places the all or part of the input graph feature data in the CPU memory needs to frequently transfer mini-batch data from CPU to GPU through a slow PCIe interconnect.
Furthermore, the minibatch method amplifies the total amount of data access because the different minibatches can have overlapping nodes.
Due to these reasons, training GNN is often throttled by CPU-GPU data transfer time.
In many of our measurements, we often find the GPU is only about 30-40\% utilized during the GNN training when the datasets do not fit in GPU memory.

To remedy this problem, in this work, we introduce \textit{Data Tiering} in GNN, which does not inflict any algorithmic changes on the training models, but yet dramatically reduces CPU-GPU data transfer volume.
Our data tiering improves GNN training in two ways.
First, it provides a statistical method by using reverse pagerank to effectively predict the importance of each node in the input graphs and identifies which nodes should be located in the GPU memory.
Second, it introduces a hardware-friendly data placement and access strategy which minimizes the cost of accessing cold data in CPU memory.
Our data placement and access strategy is quite comprehensive and it also enables more advanced optimization techniques for the multi-GPU systems with high speed GPU-to-GPU interconnects.

We evaluate our work using public frameworks PyTorch and DGL.
The demonstration of our work on realistic mini-batched training shows that our approach eliminates PCIe traffic by 87--95\% in various datasets by loading only 10\% of them into GPU memory.
With the data transfer time optimization alone from our data tiering strategy, we find the training speeds of the existing GNN implementations can be improved by 1.6--2.1$\times$.
To demonstrate the scalability of our work, we also train a dataset with 350GB of size in a system with four NVIDIA V100 32GB GPUs.
\section{Background}
\subsection{GNN and Node Classification}
\label{sec:classification}

GNNs are a series of multi-layer feedforward neural networks that propagate and transform layer-wise features following a graph structure. Among these models, a graph convolutional network (GCN)~\cite{DBLP:journals/corr/KipfW16} architecture is widely employed, which relies on the layer-wise message passing scheme. Formally, the $(l+1)$-th layer of a GNN is defined as:\footnote{We omit edge features for simplicity.}

\begin{equation} \label{eq:1}
    H^{(l+1)} = \sigma(f_w(\mathcal{G}, H^l)),
\end{equation}

\noindent where the function $f_w(\mathcal{G}, H^l)$ is determined by learnable parameters $w$ and $\sigma(\cdot)$ is an optional activation function. $\mathcal{G}$ represents a graph composed of $N$ nodes and $E$ edges.
Additionally, $H^l$ represents the embeddings of the nodes in the $l$-th layer, and $H^0$ is initialized with the nodes' input features $X$.
The node input features are stored in a $N{\times}D$ matrix where $N$ is the total number of nodes of the input graph and $D$ is the dimension of each node feature.
The size of node feature varies significantly depending on the dataset, but the typical sizes are in between 512B to 4KB.
When the node feature size is multiplied with the total number of nodes in a graph, the node feature matrix (or tensor) can reach tens of gigabytes to hundreds of gigabytes.
Therefore, storing the node feature tensors of GNN datasets is the most difficult task with the large graphs.

\begin{figure}[htbp]
\centerline{\includegraphics[width=\linewidth]{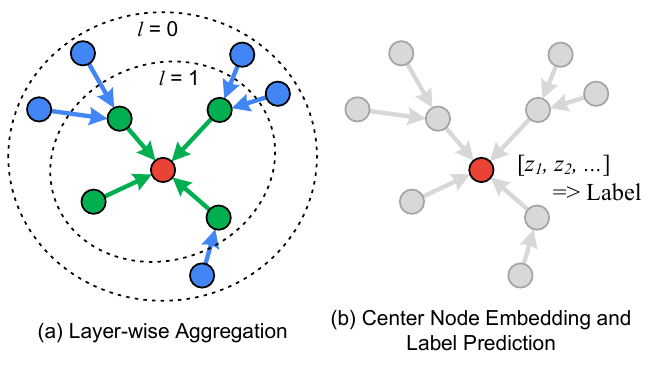}}
\caption{GNN with node feature aggregation and label prediction.}
\label{fig:preview}
\end{figure}

When $l$ is identical to the last layer of GNN, the corresponding $H^l$ tensor is the output embedding tensor $Z$.
The output embedding tensor $Z$ is used to create predicted labels and classify unclassified nodes.
For training purposes, if the nodes already have ground truth labels, then the predicted labels are compared with them to perform a backpropagation and a model update.
Similar to the image classification task, the qualities of both the trained model and the output embeddings
can benefit from using a larger dataset with more expressiveness~\cite{DBLP:journals/corr/abs-2103-09430}.

\subsection{Neighbor Sampling}
\label{sec:neighbor}

The layer-wise aggregation implementation of GNN provides a promising way of gathering relational information from graphs, but it requires reading all neighboring nodes of each layer.
With a large graph, this approach quickly shows a scalability challenge as the number of nodes that we need to read exponentially grows with an increasing number of layers.
To alleviate this scalability problem, GraphSAGE~\cite{hamilton2017inductive} introduces neighborhood sampling and aggregating approach.
By sampling a fixed number of neighboring nodes per target node instead of demanding the whole adjacency matrix, the neighborhood sampling reduces the computation and memory footprints.
With the predefined numbers of sampling per layer, we can also effectively control the size of each mini-batching in both training and inference.
%

Neighborhood sampling is applied to every neighboring node in every aggregation step.
GraphSAGE uses
a uniformly random selection process to sample the neighboring nodes to provide an enough randomness to the training process.
The commonly used hyperparameters for the neighborhood sampling size $S_{layer}$ are $(S_1, S_2) = (25, 10)$ for a 2-layer sampling approach and $(S_1, S_2, S_3) = (10, 10, 10)$ for a 3-layer approach.
It is uncommon to go beyond the three layers of sampling
because of the need to limit the size of mini-batch.
After the sampling, a sub-graph which only contains the sampled nodes is created so the computation kernel knows how to aggregate the node features of interest.
Over different iterations of training, a new sampling is done to increase the learning entropy.

When the 
node feature tensor 
is in CPU memory, the features of the sampled nodes must be transferred to the GPU memory.
Due to the slow PCIe interconnect between CPU and GPU, this data transfer process can be quite time consuming.
In this paper, we present an optimization technique called $data$ $tiering$ that exploits locality in accessing feature data and minimizes the need for cross-PCIe accesses.
\section{Data Tiering}

\subsection{Score Function}
\label{sec:score}
%
By definition, the neighbor sampling process is random and it is difficult to exactly predict which nodes will be accessed during training. Thus, we must statistically approach the problem of identifying and exploiting locality.
The first metric we can use is the out \textbf{degree} of each node in the input graph.
With a high out degree, even if the node is not selected in a specific run of neighbor sampling, the cumulative chance of the node being selected during the entire training process is higher than the less connected nodes.
Considering that we perform quite significant number of sampling per training epoch for the large graphs, this prediction is statically reasonable as we empirically prove it in Section~\ref{sec:score_vs_measure}.
In case of ogbn-papers100M, we sample about 130 millions of nodes per training epoch.

\begin{figure}[htbp]
\centerline{\includegraphics[width=\linewidth]{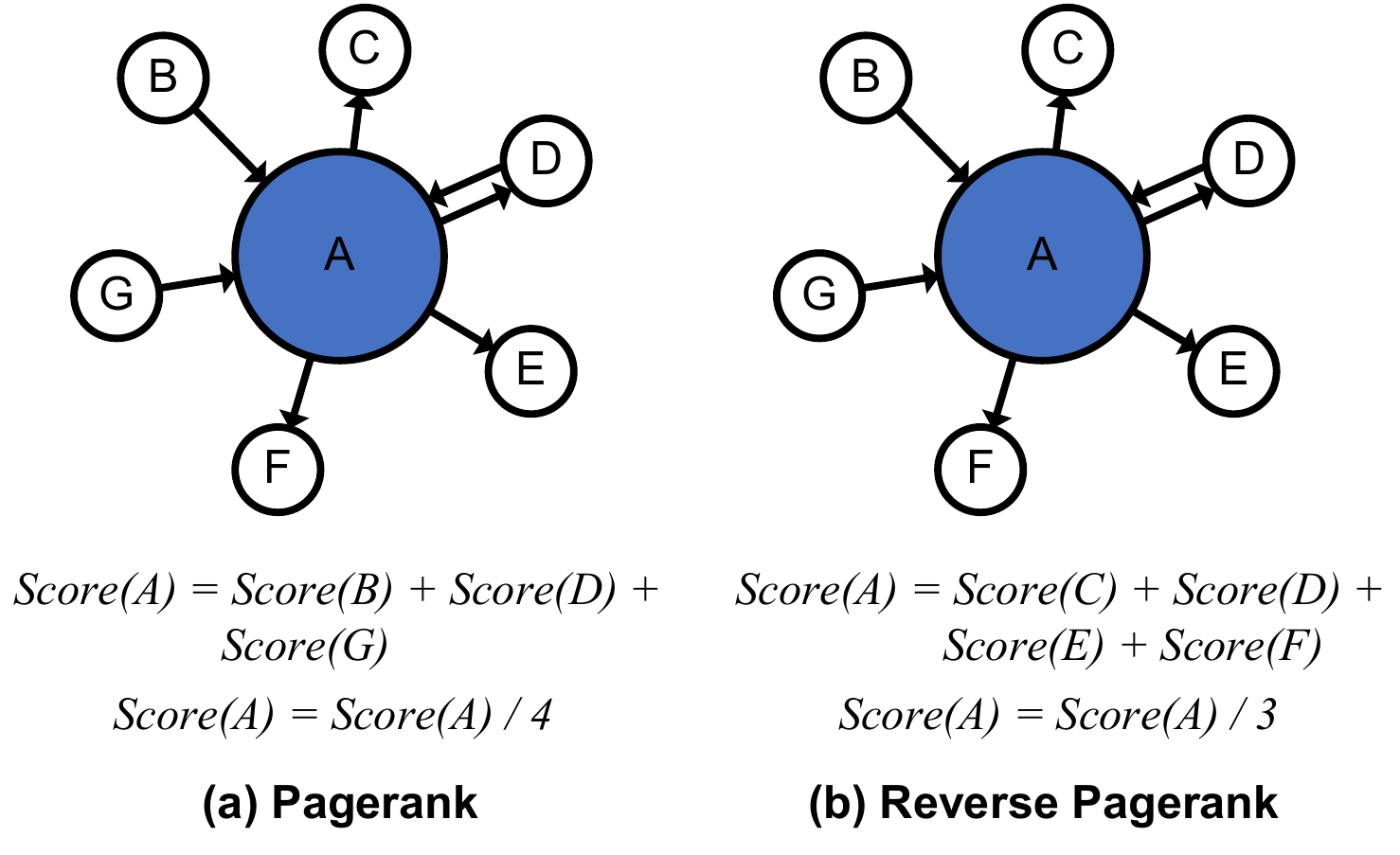}}
\caption{Snapshot of Pagerank vs. Reverse Pagerank. Only a single iteration of algorithms shown. In case of regular pagerank, the score is divided by the \textbf{out degree}, but in case of reverse pagerank, the score is divided by the \textbf{in degree}.}
\label{fig:pagerank_exp}
\end{figure}

The second option is a reverse pagerank (\textbf{R-Pagerank})~\cite{reverse_pagerank}.
In the original pagerank, the score of each node is higher if the in degree is higher and the out degree is lower.
For the reverse pagerank, it is the opposite.
In Figure~\ref{fig:pagerank_exp}, we depict the difference between the original pagerank and the reverse pagerank further.
For simplicity, we only show a case of node $A$ with single iteration, but this is done for all nodes until the score values converge in the real implementation.
In the original pagerank, to calculate the score of the source node, we sum the scores of the nodes which are targeting the source node and divide the summed score by the out degree of the source node.
Now with the reverse pagerank, we sum the scores of the nodes which are targeted by the source node and divide the summed score by the in degree of the source node.
The idea behind this mechanism is that if a certain node $A$ has many outgoing edges, it can potentially get a higher score by summing many nodes' scores.
Therefore, if there is another node $B$ which is targeting node $A$, node $B$ also gets a higher score by adding the score of node $A$.

In the context of neighbor sampling, the scoring mechanism of the reverse pagerank can be understood by referring to Figure~\ref{fig:preview}.
The green nodes are sampled while generating the embedding for the red node, referred to a node A, because they have an outgoing edge to A. The blue nodes are sampled because they have an outgoing edge to the green nodes. Therefore,  If a node can reach many nodes directly or indirectly through its outgoing edges, it is likely to be picked during the sampling process.
Since the
probability of node $A$ being picked is high, the other nodes which are targeting this node also has a relatively higher
probability of being picked when we are sampling multiple layers.
However, if the node $A$ also has a high in degree, because now there are so many nodes which can be sampled from node $A$, the other nodes should not expect the chance of them being selected too high even if node $A$ was selected.
Thus, in this case, we divide the score of node $A$ by the in degree before propagating it to the other nodes so these nodes
receive less increase to their estimated probability of being picked during sampling.
%

The potential advanatge of using the reverse pagerank over the simple degree method is to capture further multi-layer sampling patterns. 
For the simple degree method, the information we can capture is limited to a single hop of relationship, while the actual neighbor sampling can extend to multiple hops.
On the other hand, in reverse pagerank, the score value of each node is propagated to multiple layers of nodes away until the score converges to a certain limit.
Therefore, by using the reverse pagerank, it is possible to capture the subtle pattern of multi-layer sampling in the neighbor sampling in a better way.

The third option is to further incorporate 
the labeling status of the nodes 
into the reverse pagerank method.
As we explained in Section~\ref{sec:classification}, the goal of GNN training is to create a model which can
predict the labels for the
unlabeled nodes.
%
To train such models, we must be able to compare the predicted labels with the ground-truth labels. Therefore, during 
training, the nodes which we can pick to start the neighbor sampling are reduced to the nodes that come with with labels.
This means that, if we can devise a method to statistically put further emphasis to those nodes and their surrounding nodes, we can compress the search space.

Currently, the most similar existing algorithm used to achieve such goal is the personalized pagerank (PPR).
In PPR, instead of calculating the scores of all nodes in general, we select a specific node of interest and calculate the scores of the rest of the nodes from the selected node's perspective.
In other word, now the score generation process is more customized for the selected node.
However, the problem of PPR is that it only works with a single source node but not with multiple source nodes.
Indeed, running multiple separate PPR instances for multiple source nodes is algorithmically possible, but each PPR instance has $O(n)$ of space complexity where $n$ is the total number of nodes in the input graph~\cite{oraclepgx}.
Considering that we have millions of labeled nodes for any realistically large datasets, this space complexity is simply not affordable for our case.

Therefore, instead of utilizing PPR, we add some tweaks on top of the reverse pagerank algorithm by uniformly applying a weight value to the labeled nodes.
The detailed implementation of the weighting process is described in Algorithm~\ref{alg:weighted_pr}.
%
%
%
First, before we decide how to weight the labeled nodes, we need to decide how much we want to weight them.
The assumption behind the weighting is that by knowing the exact starting locations of the neighbor sampling, we can more focus on those nodes and their surroundings.
This means, that if there are few starting nodes available in the graph, the sampling tendency will be more biased toward them and their surrounding nodes.
In the opposite, if every node can be selected as a starting node, there is no starting bias and simply the nodes with high out degree is likely to be selected during the sampling.
As a result, the weighting intensity should be high if there are few labeled nodes, and the weighting intensity should be low if there are many labeled nodes.
In our algorithm, we reflect this by defining weight = (\# of all nodes) / (\# of labeled nodes).

Next, the actual weighting is done by multiplying the initial scores of the labeled nodes with the weight value we calculated (Algorithm~\ref{alg:weighted_pr}, Line \#10).
In the original pagerank algorithm, the default initial score of all nodes is 1 divided by the total number of nodes in the graph.
In general, by running pagerank long enough, the initial impact of the initial scores wear down and the scores start to converge to certain values.
To avoid this, we do not run our weighted reverse pagerank until the scores converge, but only five iterations.
The rest of the algorithm is identical to the reverse pagerank algorithm. 
We call this algorithm as weighted reverse pagerank (\textbf{Weighted R-Pagerank}).

\begin{algorithm}[tb]
   \caption{Weighted Reverse Pagerank}
   \label{alg:weighted_pr}
\begin{algorithmic}[1]
   \STATE {\bfseries Input:} graph $g$, iteration $iter$, damp $d$, \color{blue} train\_id $tid$ \color{black}
   \STATE $num\_node = num\_nodes(g)$
   \STATE $num\_train = length(tid)$
   \FOR {$i=0$ {\bfseries to} $num\_node-1$}
   \STATE $score[i] = 1 / num\_node$
   \STATE $in\_degree[i] = num\_in\_degree(g, i)$
   \ENDFOR
   \color{blue}
   \STATE $weight = num\_node / num\_train$
   \FOR {$i=0$ {\bfseries to} $num\_train-1$}
   \STATE $score[tid[i]] = score[tid[i]] * weight$
   \ENDFOR
   \color{black}
   \FOR {$i=0$ {\bfseries to} $iter-1$}
       \FOR {$j=0$ {\bfseries to} $num\_node-1$}
            \STATE $score[j] = score[j] / in\_degree[j]$
       \ENDFOR   
        \STATE $pull\_from\_neighbor(g, score)$
       \FOR {$j=0$ {\bfseries to} $num\_node-1$}
            \STATE $score[j] = (1 - d) / num\_node + d * score[j]$
       \ENDFOR         
   \ENDFOR   
\end{algorithmic}
\end{algorithm}

\subsection{Graph Node Reordering}
With the score values, now we want to split the node features into a high score group and a low score group.
The simplest way to achieve this is to reorder the node features based on the score values and divide them into top X\% of portion and bottom 100-X\% of portion.
However, at the same time, reordering the node feature tensor alone creates a discrepancy between the node IDs in the graph and the row IDs of the node feature tensor.

To resolve this, we can consider two methods.
First, we reorder the node feature tensor but not the graph nodes.
In this case, we need to create a mapping which translates the old graph node IDs to the new node feature row IDs.
Second, we reorder the node feature tensor and also the graph nodes.
In this case, we can directly use the node IDs in the graph to access the corresponding node feature in the node feature tensor.
For our implementation, we decide to use the second method because the full-scale mapping itself takes space, and with hundreds of millions of nodes, the mapping alone will be several GBs.
%

However, unfortunately, the process of reordering the graph nodes is less intuitive than reordering a 2D dense matrix like the node feature tensor because the graphs are often represented by a sparse matrix format.
Due to it's non-straightforward nature, the current implementation of graph reordering in existing frameworks like DGL has a simple sequential implementation, but this approach is too time consuming when we need to reorder a graph with hundreds of millions of nodes.
Therefore, in this work, we implement our own parallel version of algorithm to accelerate the graph {feature} reordering process.

\begin{figure}[htbp]
\centerline{\includegraphics[width=0.9\linewidth]{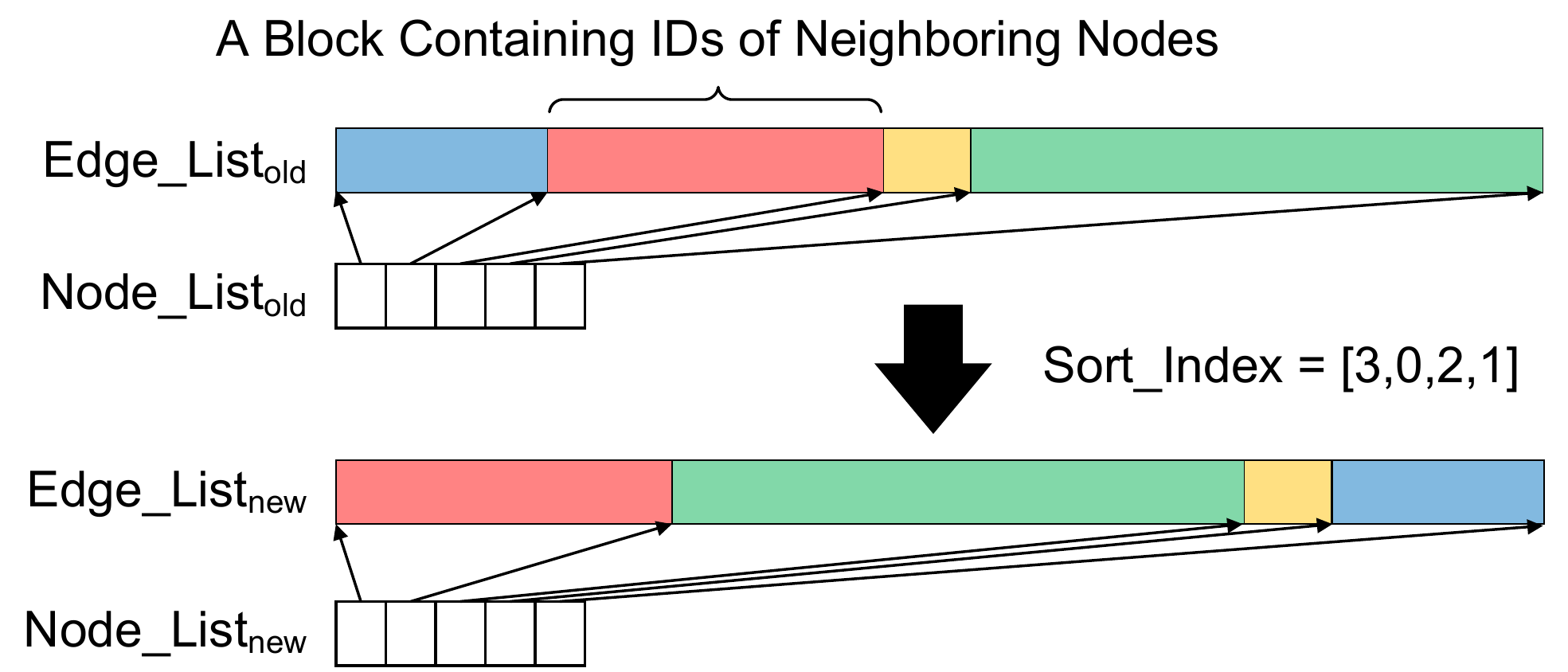}}
\caption{Graph Reordering Overview.}
\label{fig:reorder}
\end{figure}

To better understand our implementation, we first briefly explain the graph reordering problem in general.
In Figure~\ref{fig:reorder}, we show the overview of the graph reordering process.
In compressed sparse representation (CSR), the graph structure is divided into an edge list and a node list.
The edge list is a collection of many neighbor lists where each contains the IDs of nodes connected to a specific node.
To reorder a graph, we need to perform the following three tasks: First, we need to create a new node list which contains the information of new partitioning of the new edge list.
Second, we need to relocate the blocks in the edge list based on the new partitioning information.
Third, we need to update all node ID values in the edge list.

The key to parallelize the workloads is to generate a full mapping of old to new IDs in advance so the ID translation becomes a simple lookup process.
The detailed process is described in algorithm~\ref{alg:graph_reorder}.
We generate the $srt\_idx$ mapping list by creating a list of indices which can be used to sort the score values in descending order.
If the list of scores generated from Section~\ref{sec:score} is [0.1, 0.4, 0.2, 0.3], then the resulting sorted list should be [0.4, 0.3, 0.2, 0.1] and the mapping is [3, 0, 2, 1].
Now, we create a new node list $nd_{new}$ and iteratively fill the list with the sizes of the neighbor lists by using both the old node list $nd$ and the previous generated $srt\_idx$.
Since all the indices in $srt\_idx$ are unique, we can simply parallelize this iterative process.
Next, we perform a prefix sum on the $nd_{new}$ to create a new node list.

Creating a new edge list requires two steps. First, the values in the edge list should be updated and the neighbor lists in the edge list should be relocated.
To update the values, we simply index the $srt\_idx$ with the old edge list values and replace them.
This process is fully parallelizable as there is no race condition.
Next, to relocate the neighbor lists in the edge list, we use the previously created $nd_{new}$ and old $nd$ lists.
We identify the location of old neighbor list using $nd$ and place to the new location using $nd_{new}$.
This process is also easily parallelizable by distributing the neighbor list copy operation over multiple threads.
Depending on the variation of each neighbor list's length, the workload can be unbalanced between different CPU threads for this step, but we find the nodes with similar scores have similar neighbor list lengths and the workload distribution also becomes balanced.
When we define $n$ as a number of nodes and $e$ as a number edges, the total time complexity of this algorithm is either $O(nlogn)$ due to the node sorting, or $O(e)$ when the number of edges is very large.
However, thanks to our fully parallelizable approach, we find the end-to-end graph reordering takes only about 31 seconds with ogbn-papers100M dataset (Table~\ref{tab:dataset}), which has 111M nodes and 3.2B edges.

\begin{algorithm}[tb]
   \caption{Parallelizable Graph Reordering}
   \label{alg:graph_reorder}
\begin{algorithmic}
    \STATE {\bfseries Input:} graph $g$, score $score$
    \STATE $srt\_idx = indices\_of\_sorted\_list(score, descending)$
    \STATE $nd = node\_list(g), num\_node = num\_nodes(g)$
    \STATE $edg = edge\_list(g), num\_edge = num\_edges(g)$
    
    \FOR {$i=0$ {\bfseries to} $num\_node-1$}
    \STATE $neighbor\_list\_length = nd[i+1] - nd[i]$
    \STATE $nd_{new}[srt\_idx[i]+1] = neighbor\_list\_length$
    \ENDFOR
    \STATE $nd_{new}[0] = 0$
    \STATE $nd_{new} = prefix\_sum(nd_{new})$

    \FOR {$i=0$ {\bfseries to} $num\_edge-1$}
    \STATE $edg[i] = srt\_idx[edg[i]]$
    \ENDFOR

    \FOR {$i=0$ {\bfseries to} $num\_node-1$}
        \FOR {$j=0$ {\bfseries to} $nd_{new}[i+1]-nd_{new}[i]$}
            \STATE $edg_{new}[nd_{new}[i]+j] = edg[nd[i]+j]$
        \ENDFOR
    \ENDFOR
    
    \STATE \textbf{return} $(nd_{new}, edg_{new})$
    
\end{algorithmic}
\end{algorithm}

\section{Data Placement and Access}
The end goal of data tiering is to locate frequently accessed data 
in the GPU memory so we can maximize the effective bandwidth.
In this section, we describe a system level design to achieve this goal.

\subsection{Memory Allocation and Indexing Scheme}

\begin{figure}[htbp]
\centerline{\includegraphics[width=\linewidth]{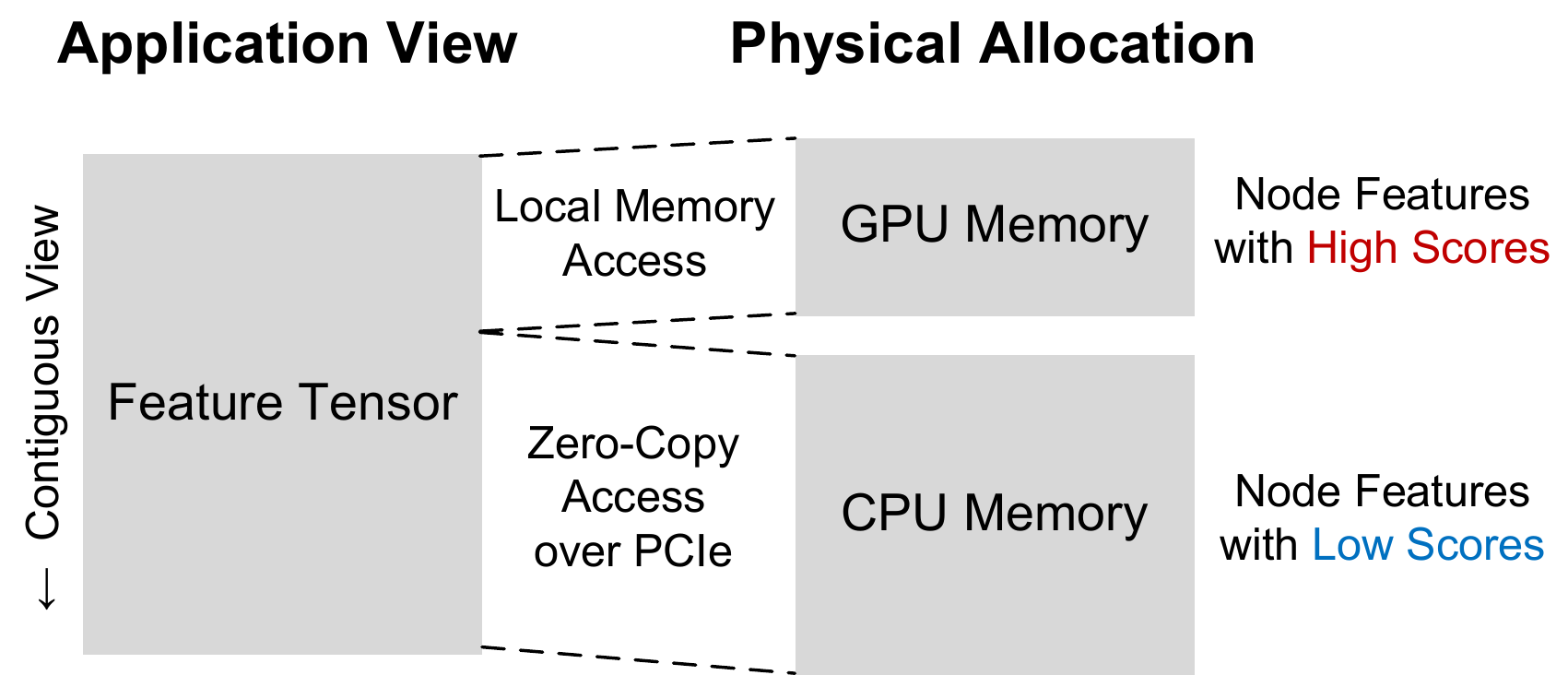}}
\caption{Simple data placement and access method overview.}
\label{fig:placement}
\end{figure}

The overall data placement and access strategy is shown in Figure~\ref{fig:placement}.
With the sorted node feature tensor, the \textit{hot data} portion with a high score is placed in the GPU memory and the rest of \textit{cold data} portion with a low score is placed in the CPU memory.
From the user application perspective, we provide a single monolithic and contiguous fake view of the two tensors so the user application can use the existing data indexing scheme.

For the cold data access, it is important to maintain a low end-to-end data transfer overhead since crossing over PCIe is already a huge burden.
One the of most common mistakes made by programmers during CPU-GPU data communications is that often the programs spend too much time on simply coordinating the data transfer.
Using cross-device data copy engines like DMA is wide spread, but it is only effective when the size of data that we want to transfer is large enough.
In the case of node feature sampling and aggregation, the size of each random access is typically between 512B and 4KB, much smaller than the size required to make DMA transfers efficient~\cite{10.1145/3297663.3310299}.
Not only that, if GPU needs to rely on CPU to initiate the data transfer as is the case of DMA, the  synchronization overhead can be completely outweigh the time cost of the transfer itself.

Therefore, in this work, we take 
a GPU-centric approach to accessing data instead of depending on DMA or CPU.
At the hardware level, the GPU-centric method is enabled by using the zero-copy access capability of NVIDIA GPUs.
The zero-copy accessible CPU memory space is mapped into the GPU page table and it allows us to directly access the CPU memory space from the GPU CUDA kernel code.
At the user (application) level,
we utilize the DGL's UnifiedTensor~\cite{10.14778/3425879.3425883} implementation to enable this data access mechanism.

With our data placement and access strategy, the overall flow of node feature access in GNN training is as follows.
First, the neighbor sampling function traverses the input graph and generates a list of node IDs.
Next, the node IDs are sent to the GPU(s).
Next, the GPU threads 
start accessing the node features with the node IDs.
While looping over the node IDs, the GPU threads check if the ID values are within the hot data boundary set by the users.
%
If the ID values are within the boundary, then the threads use a pre-stored GPU memory pointer and take the advantage of fast local memory access.
If the node ID is outside the boundary, then the threads use a pre-stored CPU memory pointer and perform the zero-copy access.

One major benefit of our approach is that we 
only attack the data transfer part of the GNN training.
The internal indexing scheme with varying GPU/CPU memory access modes of our design is completely transparent to the GNN model itself.
Such attribute makes our approach extremely modular and immediately allows the existing GNN models to be trained on large graphs.

\subsection{Tensor Distribution over Multiple GPUs}
\label{sec:multi-gpu}
\begin{figure}[htbp]
\centerline{\includegraphics[width=0.9\linewidth]{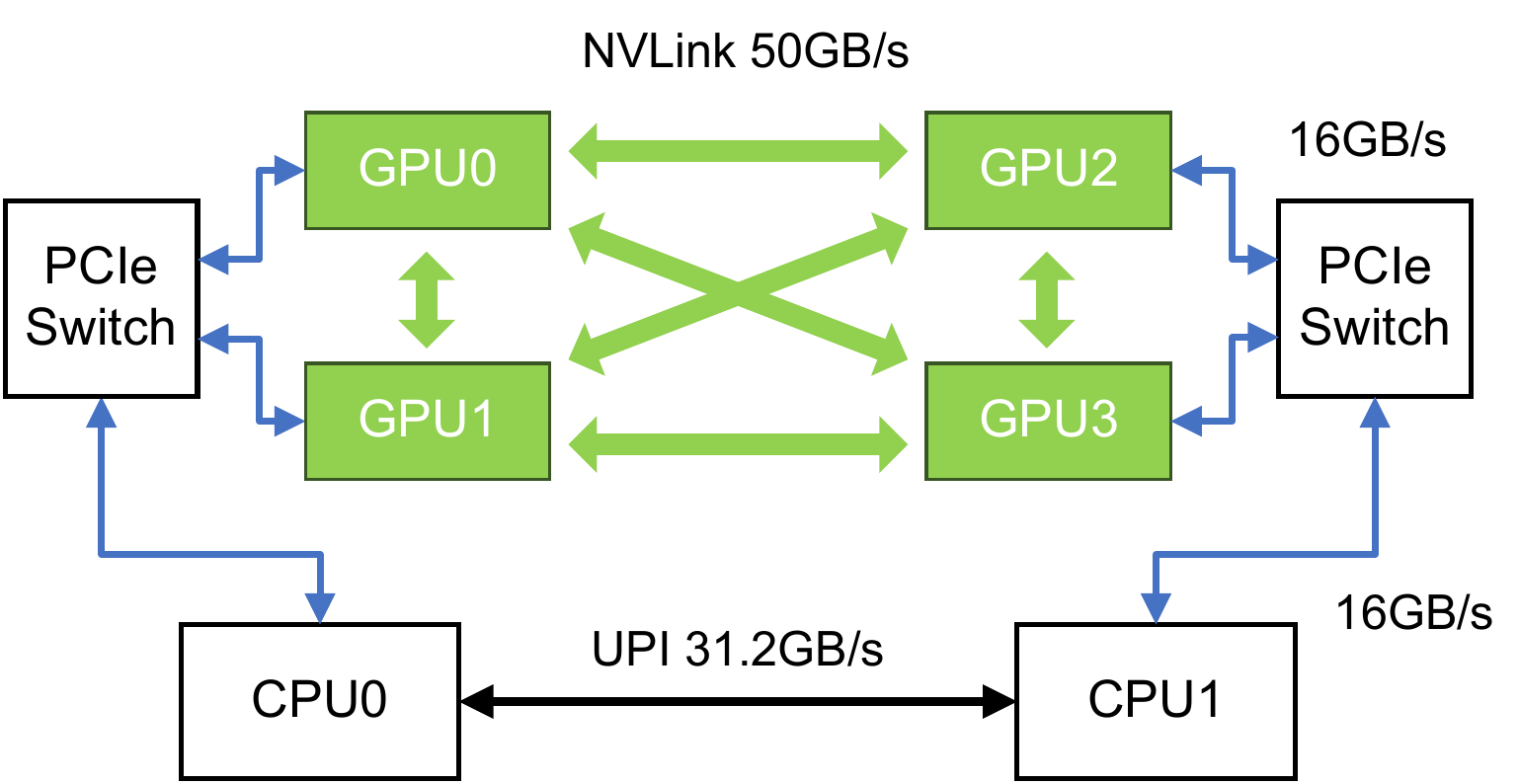}}
\caption{An example system with four NVIDIA V100 GPUs connected over NVLink. All bandwidth numbers shown are unidirectional.}
\label{fig:nvlink}
\end{figure}

\begin{figure}[htbp]
\centerline{\includegraphics[width=0.9\linewidth]{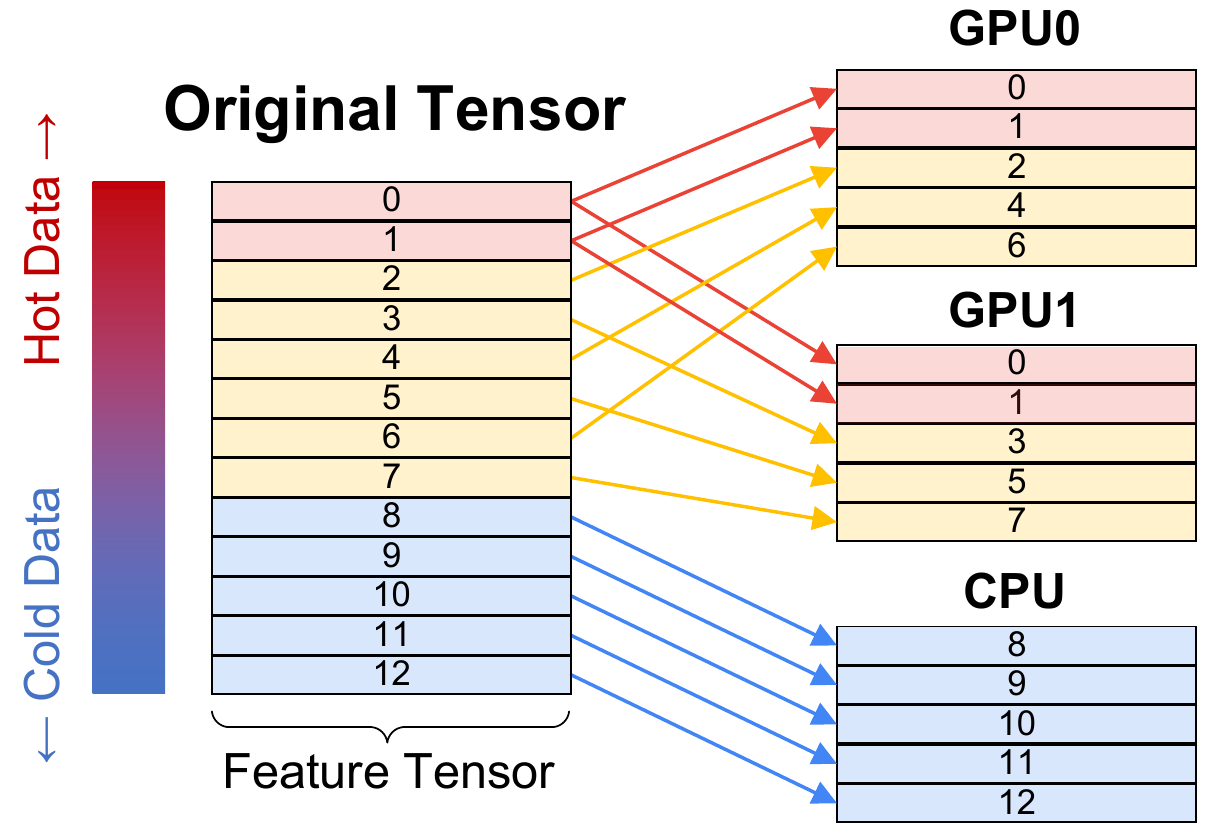}}
\caption{Data loading in multi-GPU environment.}
\label{fig:dataloading}
\end{figure}

Similar to the other 
neural network training methods, GNN training also extensively utilizes 
multiple GPUs to further accelerate the training process.
With fast GPU-to-GPU interconnects like NVLink, we can create a larger pool of collective GPU memory space (Figure~\ref{fig:nvlink}) from multi-GPU systems.
In Figure~\ref{fig:dataloading}, we show the complete view of our data tiering strategy in a multi-GPU system.
To load the hot data into this collective memory space, instead of using a na\"ive blocked partition method, we use an interleaved data loading method.
Since the node feature tensor is sorted in a descending order of the score, a simple block partitioning scheme can result in unbalanced memory and interconnect bandwidth consumption across GPUs.
With the combined GPU memory space, we can hold a larger portion of hot data in a faster tier of memory space.
The specific usage of CUDA APIs to enable our data placement strategy in multi-GPU environment is explained in Appendix~\ref{app:cuda_api}.

In our experience, the NVLink bandwidth is fast enough to hide most of the data transfer time of GNN training, but in case the data transfer time is still an issue, we can additionally replicate some hot data over multiple GPUs.
In this case, the most frequently accessed data will come from the local GPU memory, the next most frequently accessed data from the peer GPU memory, and the least frequently accessed data from the CPU memory.
The generation of the combined tensors is fully automated in our implementation.
To generate this combined tensor, users simply need to provide the sizes of local GPU tensor and multi-GPU tensor, and the number of GPUs connected over NVLink.
If there are no high speed links between GPUs, the mapping simply falls back to Figure~\ref{fig:placement}.
The optimal distribution factor of each GPU's memory capacity between the 
replicated hot data and the interleaved hot data in the collective memory space may vary depending on the dataset.
For our experiments, we simply maximize the multi-GPU tensor and do not utilize the replicated local GPU tensor.

\section{Evaluation}
\subsection{Methodology}
\subsubsection{Application \& Dataset}
For the evaluation, we implement data tiering on  PyTorch and DGL.
Since neither of the frameworks support kernel-level direct peer GPU memory access, we modify their tensor implementations to enable it.
Currently, the frameworks can only perform peer-to-peer DMAs.
We use GraphSAGE implementation of DGL to explore various neighbor sampling strategies.
For the dataset, we use the following three from Open Graph Benchmark (OGB)~\cite{hu2020ogb}: ogbn-papers100M, MAG240M, and WikiKG90M.
WikiKG90M is from a different task domain and does not come with the labels needed for node classification 
but due to the lack of public large datasets, we repurpose it as a node classification task dataset with synthetic label values.
Further details of the datasets are shown in Table~\ref{tab:dataset}.
To make our experiment realistic, we use the carefully tuned hyper parameters for different datasets which are taken from previous GNN training works with high accuracy models~\cite{DBLP:journals/corr/abs-2009-03509,paperssage,Preprint2021RUNIMPSF}.
Based on the previous works, we use (12,12,12) as neighbor sampling fanout parameter for ogbn-papers100M and (25,15) for MAG240M.
For WikiKG90M, we use the identical parameters used in ogbn-papers100M training.

\subsubsection{Hardware}
Throughout the evaluation, we use a machine with two Intel Xeon Gold 6230 CPUs and four NVIDIA V100 32GB GPUs (Figure~\ref{fig:nvlink}).
All NVIDIA V100 GPUs are connected over NVLink with 50GB/s unidirectional bandwidth per connection.
Because each GPU is connected to three other peer GPUs, the aggregated NVLink bandwidth is 3$\times$50GB/s = 150GB/s for each GPU.

\begin{table}[t]
\caption{Evaluation Datasets.}
\label{tab:dataset}
\begin{center}
\begin{small}
\begin{sc}
\begin{tabular}{lccc}
\toprule
\multirow{2}{*}{Name} & \multirow{2}{*}{\#Nodes} & \multirow{2}{*}{\#Edges} & Feature\\
&  &  & Total Size\\
\midrule
ogbn-papers100M & 111.1M & 3.2B & 53GB\\
MAG240M & 244.2M & 3.5B & 350GB\\
WikiKG90M & 87.1M & 1.0B & 125GB\\
\bottomrule
\end{tabular}
\end{sc}
\end{small}
\end{center}
\vspace{-10pt}
\end{table}

\subsection{Score Function vs. Measured Data Resue}
\label{sec:score_vs_measure}
In this experiment, we try to find out if the score functions we discussed in Section~\ref{sec:score} can correctly predict the data reusability in real GNN training.
In Figure~\ref{fig:method_comp}, we list the nodes of graphs in the X-axis in descending order of scores with the three different  scoring functions: node degree, reverse pagerank, and weighted reverse pagerank.
In the Y-axis, we show the measured access frequency of each node during GNN training, in a cumulative fashion.
In general, we find all functions can provide some level of benefits when we try to perform data tiering in GNN training.
For example, based on the scores calculated, when we keep top 10\% of nodes,  we can expect at least 35\% of hit during the training regardless of the dataset.
If we increase the ratio to 25\%, the minimum hit ratio further increases to 56\%.

Also, in general, we find it becomes easier to predict which nodes would have high data access counts if a graph has a more extreme power law distribution.
For example, WikiKG90M graph has an extremely unbalanced edge connectivity and 80\% of edges in the entire graph are connected to only 1\% of nodes.
With such extremely concentrated connections, we can observe that simply choosing a few nodes with the highest degrees automatically guarantees a very high hit ratio during the neighbor sampling.
In cases of ogbn-papers100M and MAG240M, the edge connectivities of the datasets are more balanced and the ratios are 32\% and 46\%, respectively.
However, even though the simple degree method can be effective for certain graphs, we find the weighted reverse pagerank is more preferable in general because it consistently gives the best 
prediction result.

\subsection{GNN Training Time (Single GPU)}
\label{sec:gnn_eval}

In this section, we evaluate the actual benefit of our work in GNN training.
We compared the performance of our work against the following two existing methods: 1) CPU gathering and 2) zero-copy access.
The CPU gathering method relies on CPU to gather node features and then utilize GPU DMA engine to copy the gathered node features into GPU memory.
Due to the additional data gathering process, this method not only wastes the CPU memory bandwidth, but also adds a non-negligible amount of data transfer latency to GPU.
This is the only way currently available in PyTorch to transfer scattered data in CPU memory to GPU memory.

\begin{figure}[tbp]
\centerline{\includegraphics[width=0.95\linewidth]{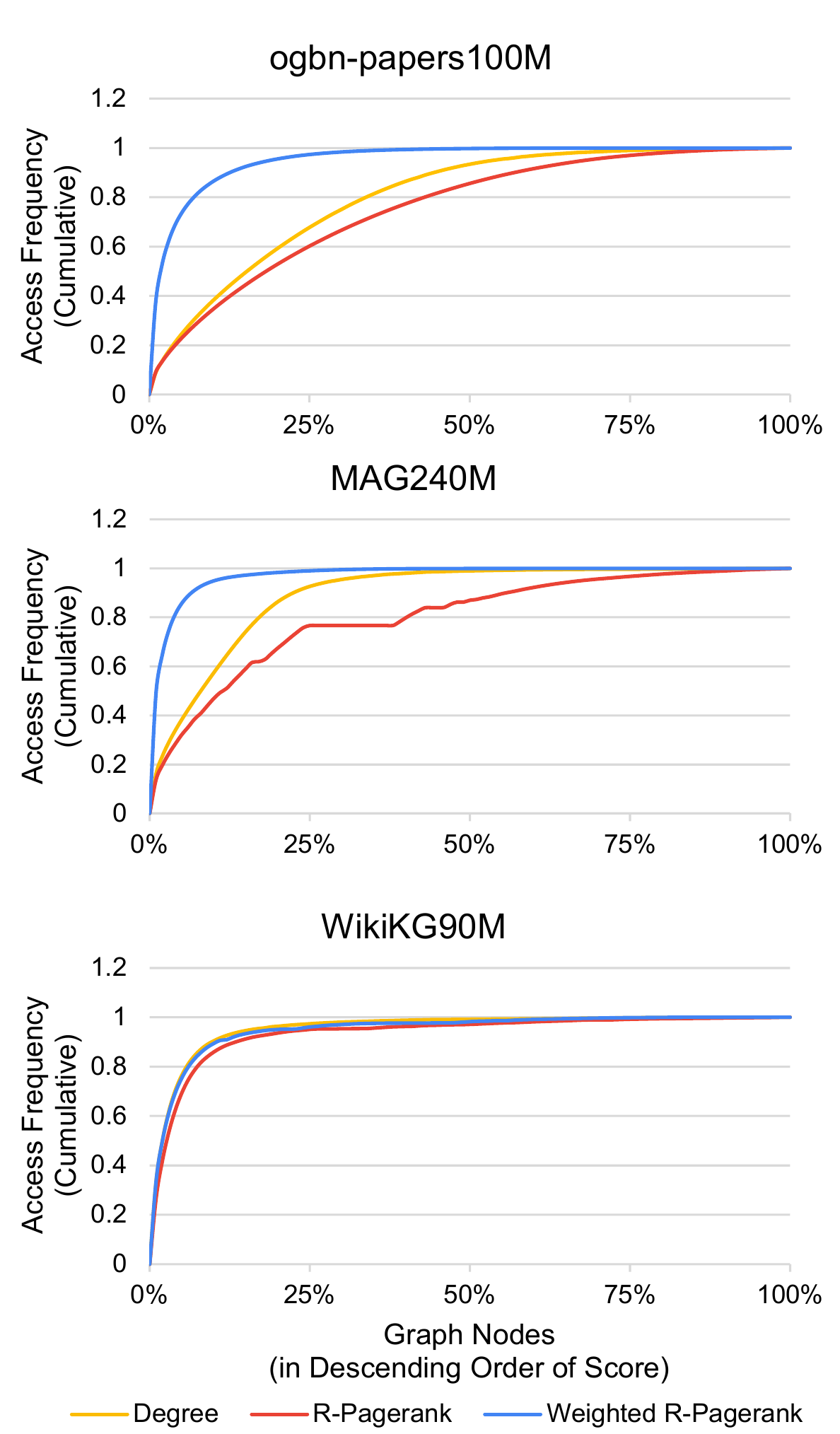}}
\caption{Access frequency distribution comparison on different datasets with different score functions. Y-axis is normalized to the total number of access.}
\label{fig:method_comp}
\vspace{-10pt}
\end{figure}

The zero-copy access method is a method recently introduced in DGL to overcome the data gathering overhead in the CPU gathering method.
With this method, GPU kernels can directly dereference CPU memory pointers and thus we do not need to rely on CPU to gather data for the GPUs.
To enable the zero-copy access capability, DGL implements a new class of tensor called UnifiedTensor which transforms the CPU tensor of PyTorch into a zero-copy accessible tensor.
In UnifiedTensor, the specific task is done by utilizing \texttt{cudaHostRegister()} API from CUDA on top of existing CPU memory allocation.
The further technical detail is identical to the process explained in Appendix~\ref{sec:user_addr}.

For our work, we first score the nodes with the weighted reverse pagerank function like in Figure~\ref{fig:method_comp}, and then reorder the node feature tensor and the graph nodes in the datasets.
For this experiment, we load 10\% of hot data for ogbn-papers100M and WikiKG90M, but only 5\% for MAG240M due to the GPU memory limitation.

\begin{figure}[tbp]
\centerline{\includegraphics[width=0.95\linewidth]{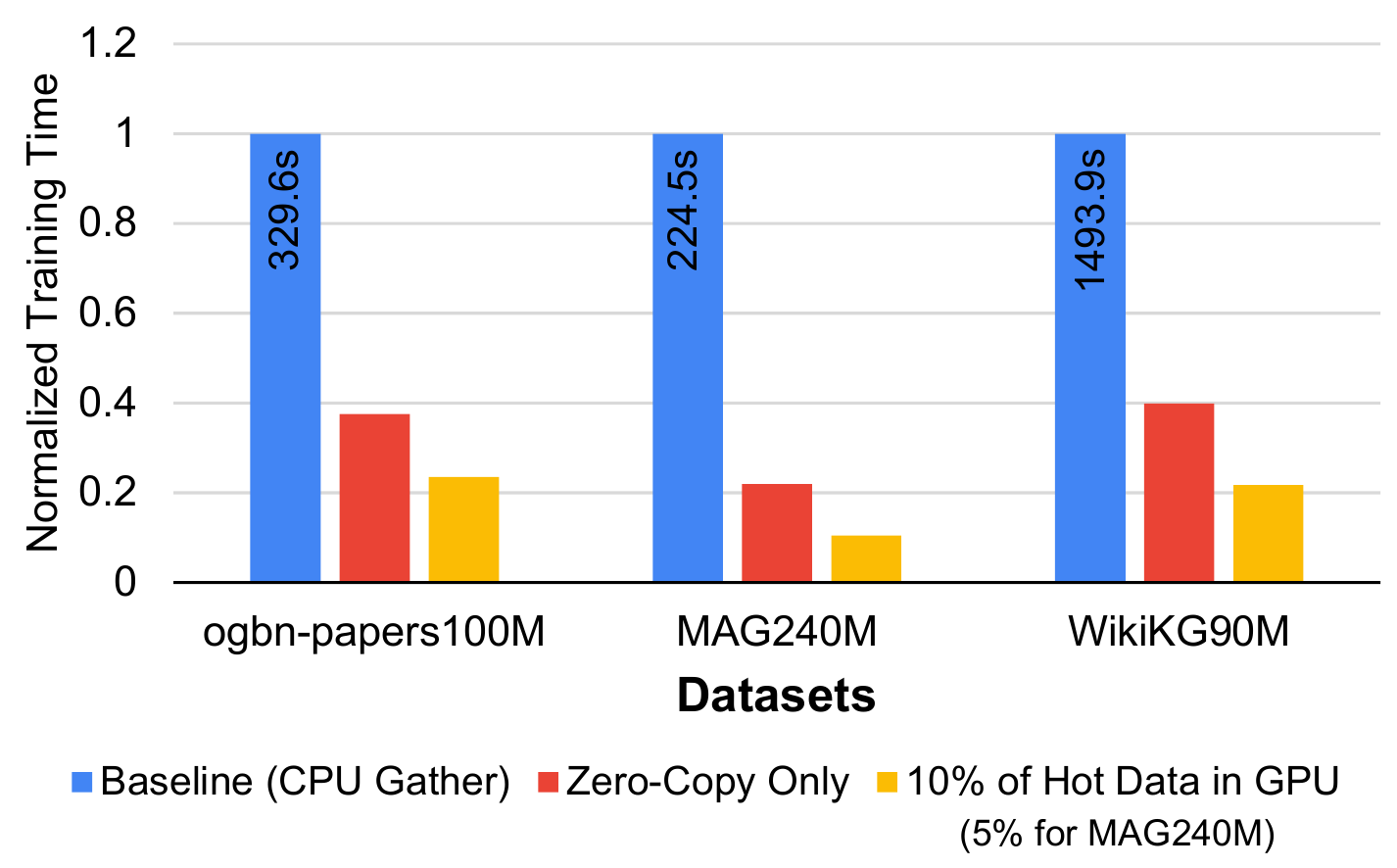}}
\caption{Single epoch training time comparison.}
\label{fig:overall_eval}
\vspace{-10pt}
\end{figure}

In Figure~\ref{fig:overall_eval}, we show the overall comparison.
From this comparison, we can first 
observe that relying on CPU to gather data results in seriously 
increasing the overall training time.
%
In this case, we find that the GPU is only about 10-30\% utilized and mostly idling.
By adopting the zero-copy access method, the training performance is visibly improved (2.5-4.6$\times$).
The zero-copy only method does not leverage any temporal data locality strategies, but simply removing CPU from the data access path significantly increases the overall performance.
Finally, with our method, the training performance is further improved by 1.6-2.1$\times$ on top of the zero-copy only method.
Considering that we have run the entire experiment on top of PyTorch and DGL with python, the benefits that we observe are immediately deliverable to the regular users as well.

\subsection{Case Study: ogbn-papers100M}
\begin{figure}[tbp]
\centerline{\includegraphics[width=\linewidth]{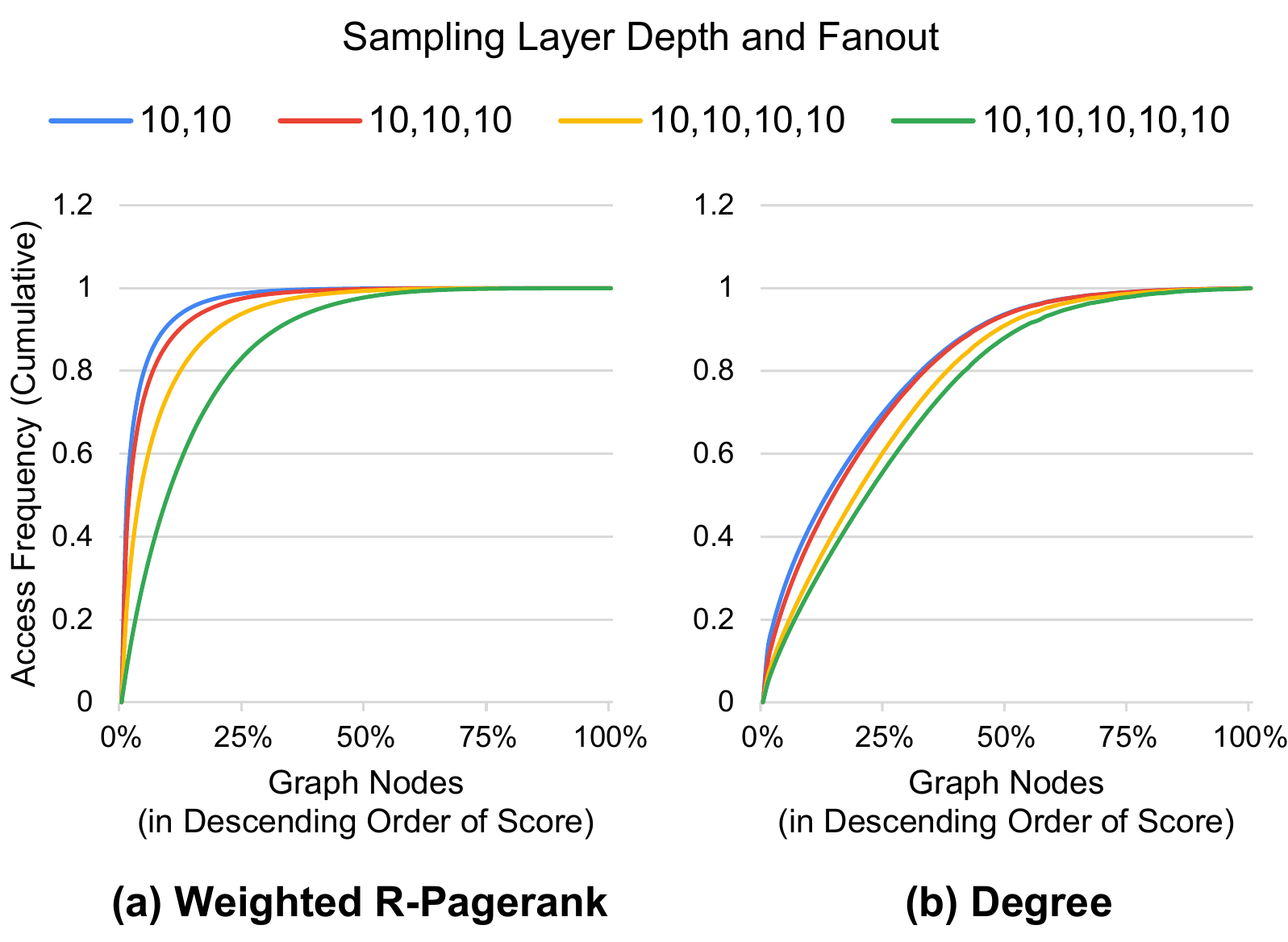}}
\caption{Access frequency distribution comparison of using different neighbor sampling parameters in ogbn-papers100M dataset.}
\label{fig:layer_comp}
\end{figure}

Now, we take ogbn-papers100M as an example 
for more detailed analysis.
First, even though we know that the most of existing GNN works use two to three layers of sampling depths, we still like to know how much increasing the sampling depth can affect the data tiering efficiency.
To understand the impact, we use different sampling depths during the GNN training and observe how the node access frequency distribution varies.
In Figure~\ref{fig:layer_comp}, we show two access frequency charts similar to Figure~\ref{fig:method_comp}, but now with the varying sampling parameters of (10,10), (10,10,10), (10,10,10,10), and (10,10,10,10,10).
For the score functions, we use the weighted reverse pagerank and the degree count.
For both cases, we can observe the accesses are now more spread out with the deeper sampling parameters.
This is an expected behavior because with a deeper sampling depth, the graph coverage of each minibatch becomes larger and also we start to access secluded nodes more frequently.

However, even with the spread out in the deeper sampling cases, we can still identify a significant portion of accesses are made to a few selected nodes.
For example, with the (10,10,10,10,10) sampling parameter, top 10\% of the highest score nodes of the weighted reverse pagerank and the degree count functions account for 52\% and 28\% of the entire accesses, respectively.
This experiment result shows that the benefit of data tiering is not immediately nullified with a growing sampling layer depth and it gives some room for the future GNN models which may attempt to sample deeper. 

\begin{figure}[tbp]
\centerline{\includegraphics[width=\linewidth]{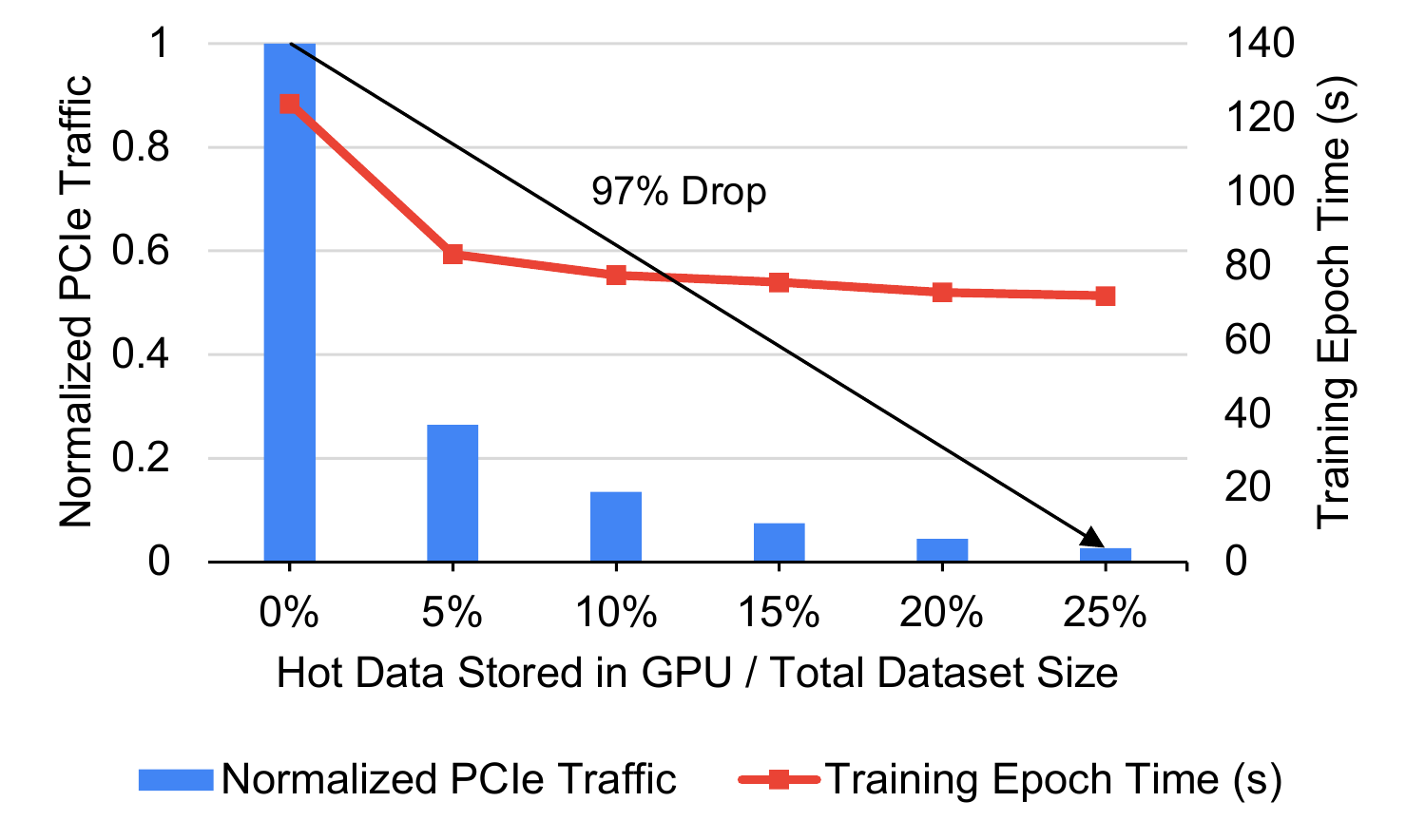}}
\caption{PCIe traffic and training time comparison in actual GNN training with increasing hot data portion loaded in GPU. ogbn-papers100M dataset used.}
\label{fig:sweep_papers}
\end{figure}

For the second analysis, we would like to more closely: 1) verify the hardware-level benefit of data tiering and 2) observe what is the impact of controlling the portion of data loaded to GPU.
To better understand these, we sweep the portion of data loaded in GPU during GNN training and measure the volume of PCIe traffic and the training time (Figure~\ref{fig:sweep_papers}).
For this experiment, we perform data tiering with the weighted reverse pagerank function on ogbn-papers100M.
To measure the PCIe traffic, we use the NVIDIA profiling tool \textit{nvprof}.
As we can see, our data placement and access strategy effectively reduces the PCIe traffic with more hot data loaded into the GPU memory.
When we compare the cases with no data loading and 25\% of data loading, we can achieve about 97\% of PCIe traffic reduction.
At this point, most of the node feature accesses are resolved within GPU and only very few data accesses need to be directed to the CPU memory over slow PCIe.

The performance gains in GNN training show a similar trend to the PCIe traffic reduction.
With the 5\% of data loaded, we can already reduce the training time by 33\% and with the 25\% of data loaded, we can further reduce the training time by 42\%.
In general, the GPU memory consumed by the training process itself is proportional to the minibatch size, and the minibatch size is exponentially proportional to the sampling depth~\cite{DBLP:journals/corr/abs-1901-00596}.
Therefore, hypothetically, if the sampling depth is very deep, the GPU memory available for hot data can be limited, but the base space complexity of GNN is relatively low.
For example, in case of ogbn-papers100M training with 3-layer sampling, we consume only about 400MB of GPU memory and the rest of the space is left unused.
Additionally, considering the trend of increasing capacity of GPU memory (e.g., NVIDIA A100 80GB) and the distributed Multi-GPU tensor solution we discussed in Section~\ref{sec:multi-gpu}, we believe the actual impact from this limitation is negligible.

\subsection{Multi-GPU Training}
\label{sec:multi_eval}
In this section, we show the performance benefit of the multi-GPU implementation of our work described in Section~\ref{sec:multi-gpu}.
In this experiment, we use four V100 32GB and therefore we can have toal 128GB of collective GPU memory space.
For the training dataset, we use MAG240M which has 350GB of node feature tensor.
For 
data placement, 
we divide the node feature tensor into two tensors, a multi-GPU tensor and a CPU tensor.
We do not allocate any space for the replicated GPU tensor.

\begin{figure}[htbp]
\centerline{\includegraphics[width=\linewidth]{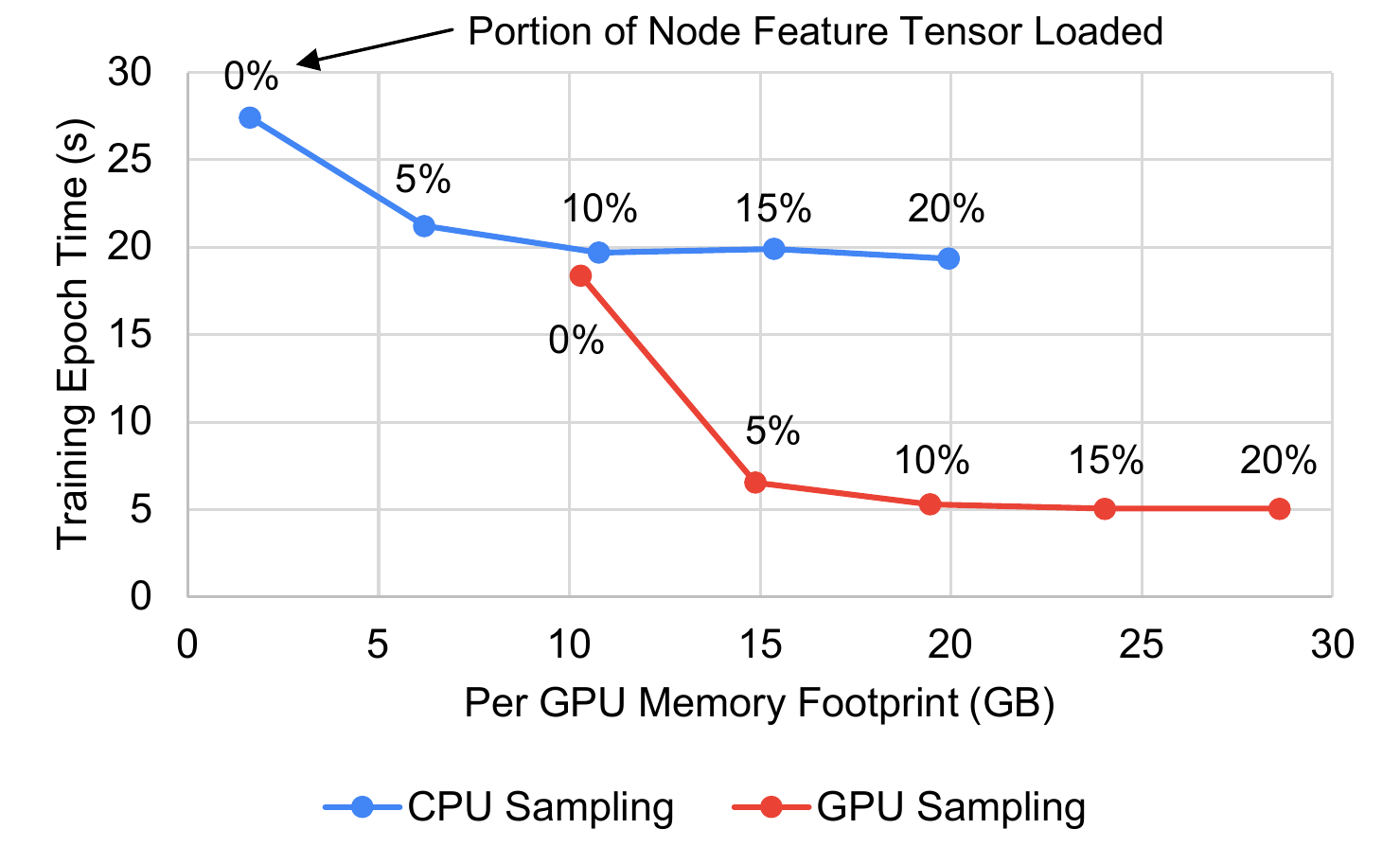}}
\caption{Multi-GPU MAG240M training time comparison while loading different amounts (in \%) of node feature tensor into GPU memory. GPU sampling requires the graph structure to be loaded in GPU memory and thus it has a higher memory footprint.}
\label{fig:mag240m}
\vspace{-10pt}
\end{figure}

In Figure~\ref{fig:mag240m}, we show the training time evaluation of MAG240M with increasing sizes of hot data loaded in the multi-GPU tensor.
Before we go into further detail of the GPU sampling results, we first focus on the CPU sampling results.
Similar to the results from Figure~\ref{fig:sweep_papers}, we observe a sharp drop of training time with 5\% of node feature loaded into GPU memory.
Beyond that, we observe only marginal performance improvements.
%
The overall performance improvements in multi-GPU training is underwhelming because the single GPU training of MAG240M in Figure~\ref{fig:overall_eval} can reach 23.5 seconds already.
This means, with four GPUs, we can reduce only about 3 seconds of training time further.

After several profiling, we find that in the multi-GPU training, the neighbor sampling process itself starts to throttle the whole training process and gives us a poor scalability.
As we described in Section~\ref{sec:neighbor}, the sampling step traverses the graph structure and generates the node IDs for a minibatch in preparation for the node feature aggregation step.
For the neighbor sampling, we have been using CPU since the graph structure is store in the CPU memory.
In a single GPU training, CPU was able to sample neighbors fast enough and provide their IDs to GPU in a reasonable amount of time.
However, now in a multiple-GPU training, the number of minibatches that we need to generate is multiplied by the number of GPUs and this starts to affect the overall training time.
Just to clarify, the implementation of CPU sampling process is already done in a parallel fashion.
In short, the amount of parallelism available from CPU is not enough to quickly traverse the graph structure and sample neighbors for multiple GPUs.
%

This problem can be overcome with the GPU-based sampling method, but only with our multi-GPU data placement strategy.
This is because to perform the GPU-based sampling, we now need to consider how to let GPUs to access the graph structure as well.
%
Of course, the simplest way of achieving this is loading the entire graph structure to each GPUs' memories, but the size of the graph structure of MAG240M alone is 30GB and it is too wasteful to load it into every GPU.
To resolve this issue, we expand the idea of multi-GPU node feature data placement strategy to the graph structure as well and distribute the graph structure over multiple GPUs.
The benefit of the combination of our data placement strategy and the GPU sampling is shown in Figure~\ref{fig:mag240m}.
When we compare the memory footprints of the CPU sampling method and the GPU sampling method, we can find in general the GPU sampling chart has been shifted to right because now the graph structure is consuming some GPU memory space.
However, in terms of overall performance, the GPU sampling method removes the CPU sampling bottleneck and notably increases the training speed in the multi-GPU setup.

As we can see, even though we initially designed our data placement and access methodology mainly for the node feature tensor, it can be expanded to different types of data structures as well.
We believe the other DNNs which have sparse data access patterns like what we observe with the graphs would benefit the most from our design.

\section{Related Work}

GNS~\cite{gns_kdd21} samples a global cache of nodes periodically for all mini-batches and stores them in GPUs. It 
employs a 
preferential sampling approach in generating mini-batches, which gives 
priority to
neighbors that exist in the GPU cache. This can greatly reduce data copy between CPU and GPU, but it requires the modification of native node-wise sampling algorithm, which cannot be easily extended to other sampling methods.
On the other hand, the design of our data tiering is orthogonal to these sampling methods, which makes it sampling-agnostic.
%
Furthermore, the method in GNS lacks the capability to leverage multi-GPU memory to store the cached data.
%

LazyGCN~\cite{ramezani2020gcn} periodically samples mega-batches and recycles the sampled nodes within a mega-batch to generate mini-batches. This reduces the overhead of data preparation. Though, LazyGCN is sampling-agnostic, it requires a large mega-batch size, regardless of which sampling method is used, to guarantee the model accuracy. 
With node-wise neighbor sampling, it can easily run out of GPU memory on large graph such as ogbn-papers100M.

PaGraph~\cite{10.1145/3419111.3421281} and AliGraph~\cite{zhu2019aligraph} provide static node caching scheme in GNN training, but their caching strategies are simply limited to high out degree nodes and their entire cache managements are done by CPU.
This approach makes GPUs to always synchronize with CPU to access data.

\section{Conclusion}
In this work, we presented a data tiering technique for GNN training.
In general, we find the training time of GNN can be easily improved with well-defined data placement and rearrangement optimizations.
Our data tiering strategy is a novel solution that does not affect the algorithm of GNN at all but still maximizes the benefit of multi-tier memory subsystem of modern hardware.
We further demonstrate that our approach improves the scalability of multi-GPU training.
We demonstrated our work by using existing libraries such as PyTorch and DGL, and our data tiering implementation can be immediately adopted by the end users.


\bibliography{example_paper}
\bibliographystyle{mlsys2022}

\clearpage
\appendix
\section{Memory Allocation With CUDA}
\label{app:cuda_api}

In this section, we explain some of the technical challenges of creating shared address space in python program and describe how to expose certain device's memory space to GPU by using several CUDA APIs.

\subsection{Unified Virtual Memory (UVM)}
By default, CUDA provides a unified memory addressing scheme for GPU programming with the unified virtual memory (UVM) implementation.
This capability is backed by the memory managed unit of GPU which allows GPU to map any system addresses into its virtual address space.
Therefore, with this capability, it is possible to weave multiple separate physical addresses into a single contiguous virtual address space.
The underlying idea of this virtual addressing in GPU is very similar to the CPU virtual addressing which allows programs to observe physically non-contiguous space like a contiguous space.
The actual allocation of this memory space can be done by calling \texttt{cudaMallocManaged()}.
By default, this API only creates a shared virtual address space of GPUs and the actual physical allocation can be managed by the programmers.
By using several memory hints with \texttt{cudaMemAdvise()}, the programmers can control the actual location of data in a GPU page granularity (64KB).
For example, we can first allocate 1GB space of memory with \texttt{cudaMallocManaged()} and assign first 512MB to CPU and the latter 512MB to certain GPU's memory.
After the allocation, when we need to access certain piece of data in the UVM space, GPU automatically generates the necessary type of memory request depending on the page table entry (e.g., local memory access for data in GPU and PCIe memory access for remote access).

However, there is one major drawback of this method with most of the python-based DNN frameworks.
Due to the global interpreter lock (GIL) implemented in python, only one thread in a process can actually proceed in python code.
To avoid this issue, the programmers either need to 1) add a C/C++ module which can run in multithreaded manner or 2) launch multiple processes instead of multiple threads to run python codes in parallel.
In case of PyTorch, it takes the second method~\cite{DBLP:journals/corr/abs-2006-15704} for multi-GPU training.
With the multiprocessing method, the overall workload is as follows.
First, the master process setup some environment.
Second, the master process spawns multiple child processes.
Each child process take control of each GPU.
When there are 8 GPUs, then we need 8 child processes.
Next, each child process train neural network on their own, and occasionally synchronize together to update their gradients.

This approach greatly simplifies the distributed training model in python, but the multiprocess also isolates the virtual address spaces between different processes.
Therefore, in case of CPU memory, different processes cannot directly access other processes' memory space, but they must first create a \textit{shared memory} space.
Only after that, the data placed in the shared memory space can be directly accessed across different processes.
The allocation of the shared memory space can be done by the APIs existing in operating systems such as Linux.

Unfortunately, for CUDA, there is no such single universal method which allows all GPUs running on different processes to share a single virtual address space.
The scope of memory allocated by \texttt{cudaMallocManaged()} is limited to a single process.
Therefore, the implementation of UVM is extremely convenient, but it is not suitable under the python environment.
There has been a several years of effort to remove GIL from python for better multithreading implementations, but it is unlikely to be done in any foreseeable future from now.
However, hypothetically, if the GIL is removed and if we can freely use multithreading instead of the multiprocessing in python, the implementation of our data tiering can be radically simplified by using UVM.

\subsection{User-Managed Virtual Address}
\label{sec:user_addr}

The alternative method for CUDA is to generate mappings for all individual allocations and manage them manually.
For example, in case of Figure~\ref{fig:placement}, we need to keep two pointers: one local GPU pointer and one remote CPU pointer.
In case of multi-GPU environment, each GPU needs to keep: its own GPU memory pointer, peer GPUs' memory pointers, and one CPU memory pointer.
This can be considered as a software-managed or user-managed virtual addressing.

\begin{figure}[htbp]
\centerline{\includegraphics[width=\linewidth]{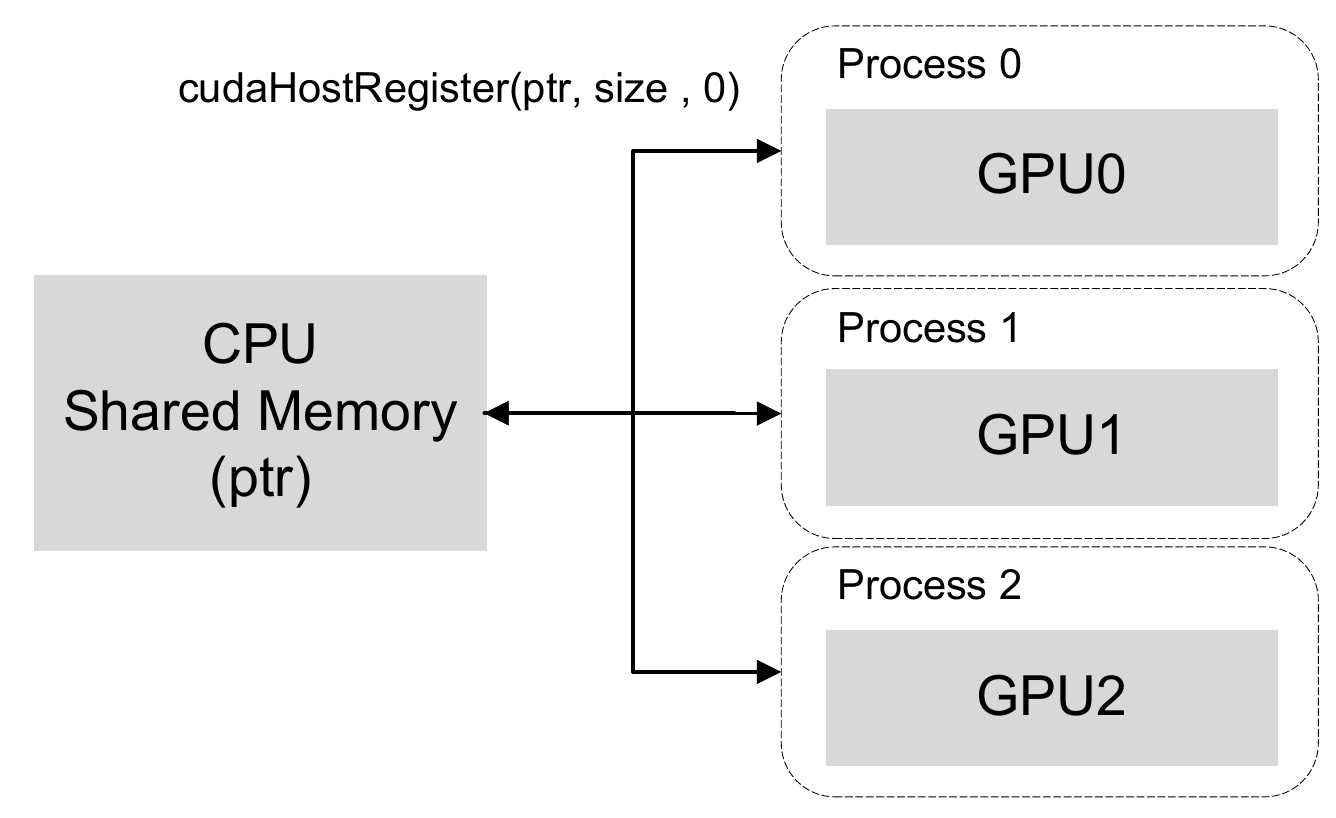}}
\caption{cudaHostRegister() is used to map memory space allocated in CPU into GPU. cudaHostRegister() needs to be called by all processes individually.}
\label{fig:cpu_mem}
\end{figure}

To map the CPU memory space into GPU memory space, we use \texttt{cudaHostRegister()} API (Figure~\ref{fig:cpu_mem}).
This API does not newly allocate a space in CPU, but it maps the memory already allocated in CPU into the GPU memory space.
To allow multiple GPUs running in multiple different processes to access the same data in CPU, we allocate the shared memory space in CPU and then let all processes to call \texttt{cudaHostRegister()} on the shared memory space.
A memory space allocated by other memory allocator such as \texttt{malloc()} cannot be shared with other processes, and therefore we must use the shared memory space in CPU.

\begin{figure}[htbp]
\centerline{\includegraphics[width=\linewidth]{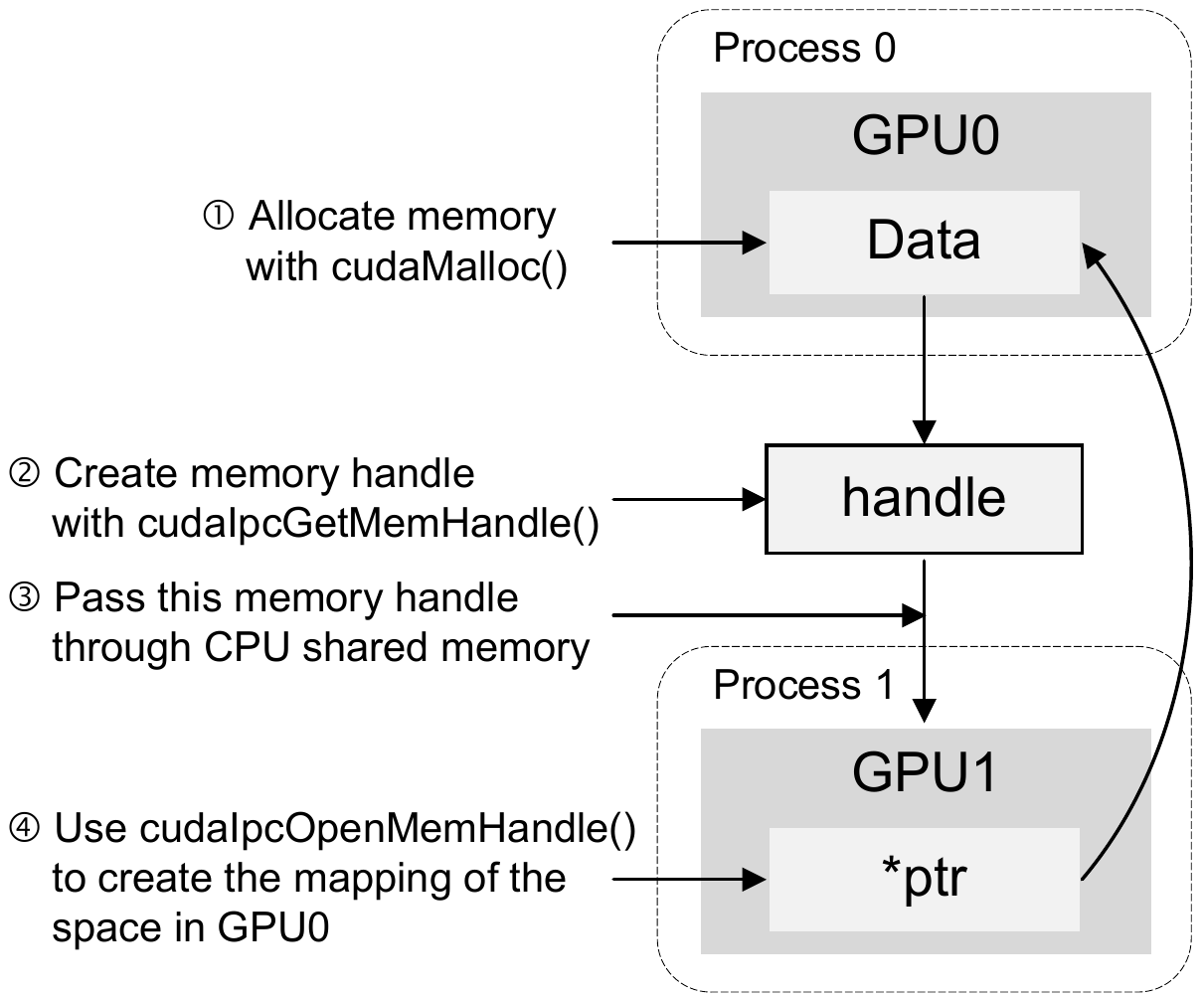}}
\caption{Peer-to-Peer GPU memory sharing mechanism in multiprocessing setup.}
\label{fig:ipc_mem}
\end{figure}

For the GPU memory sharing, we use \texttt{cudaMalloc()}, \texttt{cudaIpcGetMemhandle()}, and \texttt{cudaIpcOpenMemhandl()} APIs.
\texttt{cudaMalloc()} allocates a memory space in GPU and \texttt{cudaIpcGetMemhandle()} creates a memory handle which can be shared with other process to create a virtual mapping of the originally allocated GPU space.
In short, this \textit{memory handle} can be understood as a medium to share the virtual mapping between two different processes.
\texttt{cudaIpcOpenMemhandl()} takes the memory handle created by \texttt{cudaIpcGetMemhandle()} and maps the other GPU's memory space into the GPU device which belongs to the current process.
The overview of this process is shown in Figure~\ref{fig:ipc_mem}.

\subsection{Combined Tensor}
The combined tensor we proposed in Section~\ref{sec:multi-gpu}, is basically the table which contains the pointers of different memory allocations in different devices (e.g., CPU and peer GPUs).
Depending on the index value from the user space, we pick the corresponding pointer and let the GPU kernel to access the node feature.
The brief mechanism of this process is explained in Listing~\ref{lst:idx}.
The real CUDA kernel implementation of the indexing function is more complicated than the code provided here for several optimization purposes.

\begin{minipage}{\linewidth}
\begin{lstlisting}[language=C++, caption=Indexing the Combined Tensor., label=lst:idx]
#define WARP_SIZE 32
void index_feature(int row_id, int **table, int *threshold, int feat_len, int gpu_num, int *output_feat) {
  int tid = blockIdx.x * blockDim.x + threadIdx.x;
  int *row_ptr;
  local_boundary = threshold[0];
  multi_boundary = threshold[1];
    
  // Data is in local GPU tensor
  if (row_ids[i] < local_boundary) {
    row_ptr = table[0] + feat_len * row_ids[i];
  }
  // Data is in multi-GPU tensor
  else if (row_ids[i] < multi_boundary) {
    int offset = row_ids[i] - local_boundary;
    row_ptr = table[offset % gpu_num + 1] + feat_len * offset / gpu_num;
  }
  // Data is in CPU tensor
  else {
    int offset = row_ids[i] - multi_boundary;
    row_ptr = table[gpu_num + 1] + feat_len * offset; 
  }
    
  for (int i = tid; i < feat_len; i+=WARP_SIZE) {
    output_feat[i] = row_ptr[i];
  }
}
\end{lstlisting}
\end{minipage}


\end{document}